\documentclass[11pt]{article}

\usepackage[a4paper, left=3.5cm,top=3cm,right=3.5cm,bottom=3.6cm]{geometry}
\usepackage{boldline}
\usepackage{concmath}
\usepackage[T1]{fontenc}

\usepackage[utf8]{inputenc}

\usepackage{authblk}
\usepackage[utf8]{inputenc}
\usepackage[T1]{fontenc}
\usepackage{lmodern}
\usepackage{epstopdf}
\usepackage[english]{babel}
\usepackage{amsmath,amsthm,amssymb}
\usepackage{dsfont}
\usepackage{mathtools}
\usepackage{subcaption}
\usepackage[color=green!40]{todonotes}
\usepackage{bbm}
\usepackage{tikz}

\usetikzlibrary{matrix,chains,positioning,decorations.pathreplacing,arrows}

\DeclarePairedDelimiter{\abs}{\lvert}{\rvert}

\DeclareGraphicsExtensions{.eps,.pdf,.png,.jpg}

\numberwithin{equation}{section}
\numberwithin{figure}{section}
\numberwithin{theorem}{section}

\usepackage{cite}
\usepackage{siunitx}

\author{Nadja Gruber}
\author{Stephan Antholzer}
 
\affil{Department of Mathematics, 
University of Innsbruck\authorcr
Technikerstra{\ss}e 13, 6020 Innsbruck, Austria\authorcr
{\tt nadja.gruber@uibk.ac.at}
 }

\author{Werner Jaschke}
\author{Christian Kremser}
 
\affil{Department of Radiology\authorcr
Medical University of Innsbruck\authorcr
Anichststra{\ss}e 35, 6020 Innsbruck}

\author{Markus Haltmeier}

\affil{Department of Mathematics, 
University of Innsbruck\authorcr
Technikerstra{\ss}e 13, 6020 Innsbruck, Austria\authorcr
 {\tt markus.haltmeier@uibk.ac.at}
 }

\title{A Joint Deep Learning Approach for Automated Liver and Tumor Segmentation}

\date{Version 2 (original submission: February 20, 2019)}

\begin{document}
\maketitle

\begin{abstract}
Hepatocellular carcinoma (HCC) is the most common type of primary liver cancer in adults, and the most common cause of death of  people suffering from cirrhosis. The segmentation of liver lesions in CT images allows assessment of tumor load, treatment planning, prognosis and monitoring of treatment response. Manual segmentation is a very time-consuming task and in many cases, prone to  inaccuracies and automatic tools for tumor detection and segmentation are desirable.  In this paper, we compare two network architectures, one that is composed of one neural network and manages the segmentation task in one step and one that consists of
 two consecutive  fully convolutional neural networks. The first network segments 
 the liver whereas the second network segments the actual tumor inside the liver.  
 Our  networks are trained on a subset of the LiTS (Liver Tumor Segmentation) Challenge and evaluated on data provided from the radiological center in Innsbruck. 
\end{abstract}

\section{Introduction}

Liver cancer remains associated with a high mortality rate, in part related to initial diagnosis at an advanced stage of disease. Prospects can be significantly improved by earlier treatment beginning, and  analysis of CT  images is a main diagnostic tool   for early detection of  liver tumors.  Manual inspection  and  segmentation is a labor- and time-intensive process yielding relatively imprecise results in many cases. Thus, there is interest in developing automated strategies to aid in the early detection of lesions. 
Due to complex backgrounds, significant variations in location, shape and intensity across different patients, both, the automated liver segmentation and the further detection of tumors, remain challenging tasks. 

Semantic segmentation of CT images has been an active area of research over the past few years. Recent developments of deep learning have dramatically improved the performance of artificial intelligence. Deep learning algorithms, especially deep convolutional neural networks (CNN) have considerably outperformed their competitors in medical imaging. One of the most successful  CNNs architectures is the s-called
U-Net \cite{ronneberger2015u}, which has won several competitions in the field of 
biomedical image segmentation.

\begin{figure}[htb!]

\centering
\resizebox{0.99\linewidth}{!}{
\begin{tikzpicture}

\node[inner sep=0pt] (russell) at (-2.8,4.5)
    {\includegraphics[width=.3\textwidth]{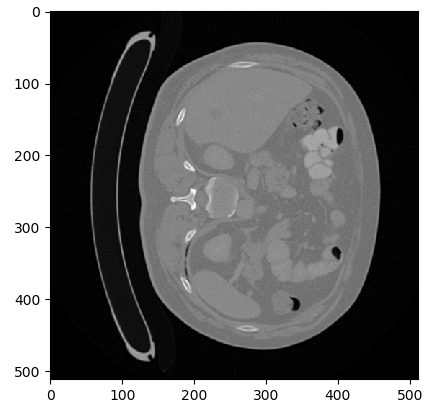}};

\draw[->] [very thick, gray](0.5, 4.5)-- (1,4.5);

\draw[->] [very thick, gray](15, 4.5)-- (15.5,4.5);

\draw[-] [very thick, gray](18.5, 4.5)-- (19.5,4.5);
\draw[-] [very thick, gray](19.5, 4.5)-- (19.5,-1);
\draw[-] [very thick, gray](19.5, -1)-- (-2.5,-1);
\draw[->] [very thick, gray](-2.5, -1)-- (-2.5,-3.5);

\node[inner sep=0pt] (russell) at (-5,-13)
    {\includegraphics[width=.3\textwidth]{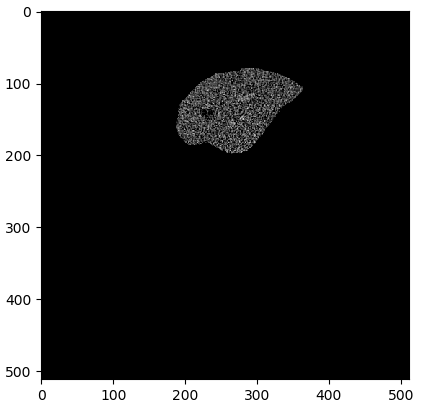}};

\draw[fill=black](-1.4, -8)circle(1pt);
\draw[->] [very thick, gray](0.5, -7)-- (1,-7);

\draw[fill=gray](8.5, -7.5)circle(0.000000007pt)node[above, gray]{\text{(B) }};
\node[inner sep=0pt] (russell) at (8.0,4.5)
    {\includegraphics[width=0.7\textwidth]{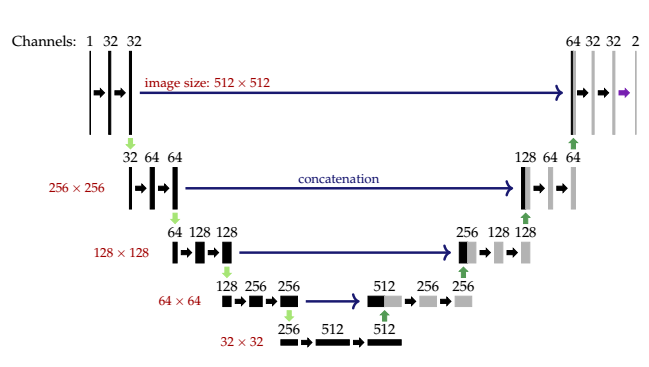}};
\node[inner sep=0pt] (russell) at (8.0,-7.5)
    {\includegraphics[width=0.7\textwidth]{miniunet.png}};
\draw[->] [very thick, gray](15,-7)-- (15.5,-7);
\node[inner sep=0pt] (russell) at (-8, -7)
    {\includegraphics[width=.3\textwidth]{train1.png}};
\node[inner sep=0pt] (russell) at (-2.5, -7)
    {\includegraphics[width=.3\textwidth]{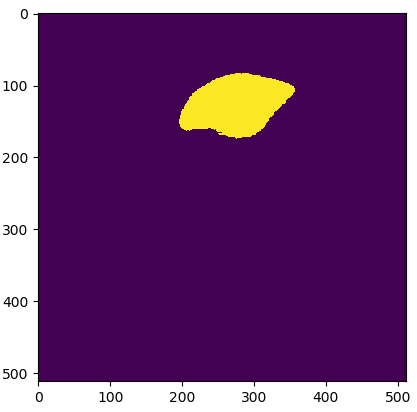}};
\node[inner sep=0pt] (russell) at (18.5,4.5)
    {\includegraphics[width=.3\textwidth]{liver.png}};
\node[inner sep=0pt] (russell) at (18.5,-7)
    {\includegraphics[width=.3\textwidth]{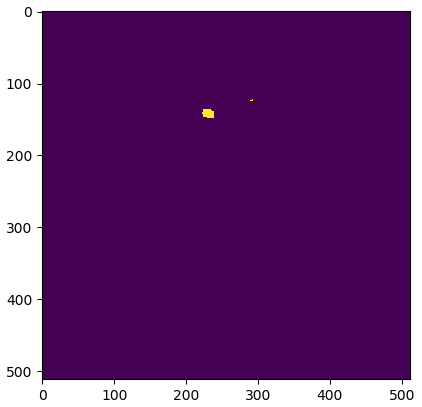}};    
    \draw[fill=black](-5.25, -7)circle(0.7pt);
   \draw [decorate,decoration={brace,mirror,amplitude=10pt}]
    (-8.5,-10) -- (-2,-10) ;
\end{tikzpicture}
}

\caption{\textbf{Illustration of Network architecture for automated semantic liver and tumor segmentation} that yielded the most promising results. The model consists of two sequential  U-Nets. 
The raw images are fed into the first network, and 
the output is a binary image. The original image  multiplied by the obtained liver mask represents the input of the second U-Net. The final output is a binary 
image in which label $1$ is assigned to tumor.} \label{fig:model}
\end{figure}

We investigate a deep learning  strategy  that jointly segments the liver and the 
lesions in CT images. As in  \cite{christ2017automatic}, we use a network 
formed of two consecutive  U-Nets.  The first network  performs liver segmentation,
while the second one incorporates the output of the first network and   segments 
the lesion. We propose a  joint  weighted loss function combining  
the outputs of both networks. The   network   is trained on a subset of the LiTS (Liver Tumor Segmentation Challenge) and evaluated on different data collected at the 
radiological center in Innsbruck. For  our initial   experiments, we perform 
 consecutive training, with which we already obtain quite accurate results.

\section{Joint Deep Learning Approach}

The overall architecture of the model yielding more accurate results is illustrated in Figure\ref{fig:model}. For the semantic segmentation task of liver and tumor, we propose a model consisting of two modified U-Nets. Related FCN architectures have been proposed in   \cite{vorontsov2018liver,han2017automatic,chlebus2018deep,christ2017automatic}. 

\subsection{One-Step U-Net Architecture}\label{sec:one}

\begin{figure}[htb!]

\centering
\resizebox{1\linewidth}{!}{
\begin{tikzpicture}

\node[inner sep=0pt] (russell) at (-2.8,4.5)
    {\includegraphics[width=.25\textwidth]{train1.png}};

\draw[->] [very thick, gray](0, 4.5)-- (0.3,4.5);

\node[inner sep=0pt] (russell) at (6.5,4.5)
    {\includegraphics[width=0.55\textwidth]{miniunet.png}};
\draw[->] [very thick, gray](12.8, 4.5)-- (13.1,4.5);
\node[inner sep=0pt] (russell) at (16.6,4.5)
    {\includegraphics[width=0.25\textwidth]{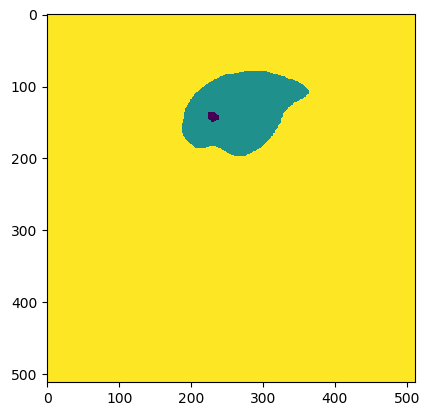}};
\end{tikzpicture}
}
\caption{Illustration of network architecture for automated liver and tumor segmentation executed in one step. The final output is a discrete class label output in which the tumor has value $1$, liver $0.5$ and background $0$.}
\label{fig:joint}
\end{figure}

As a first attempt we apply a very intuitively and straight-forward workflow which is visualized in Figure~\ref{fig:joint}. The preprocessed images are fed into the network. The depicted segmentation problem can be regarded as multi-class label classification whereas each pixel must be assigned a certain probability of belonging to class tumor, liver or other tissue. Since U-Net is known to handle semantic segmentation tasks of medical images very well, we decided to start with this quite obvious approach.
 In contrast to the mathematical modelling (see~\ref{sec:general}) of the deep learning approach illustrated in~\ref{fig:model} we generate $3$ segmentation masks $T_k, L_k, O_k, k = 1, \dots, N$ representing binary images, where class label $1$ stands for tumor, liver and other tissue, respectively. The goal is to find a parameter set $\xi$ such that the network
\begin{equation*}
\mathbb{C}_{\xi}\colon \mathbb{R}^{512 \times 512} 
 \rightarrow \left[0,1\right]^{512 \times 512 \times 3}
\end{equation*}
fulfils $\mathbb{C}_{\xi}(X_k)_{_0}\simeq T_k$, $\mathbb{C}_{\xi}(X_k)_{_1}\simeq L_k$ and $\mathbb{C}_{\xi}(X_k)_{_2}\simeq O_k$.

\subsection{Sequential U-Net Architecture}\label{sec:seq}

\begin{figure}[htb!]
\begin{center}
				
\begin{subfigure}[b]{0.8\textwidth}				
  	\includegraphics[width=0.48\textwidth]{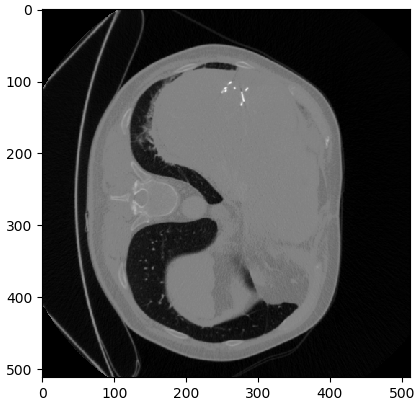}
  	\hfill 
  	\includegraphics[width=0.48\textwidth]{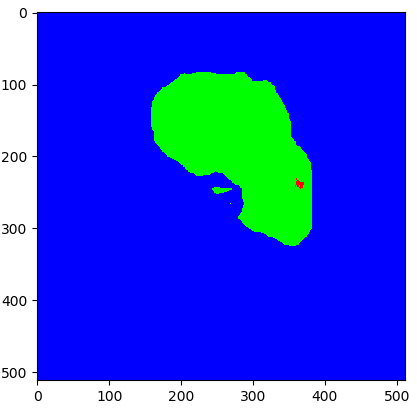}  	
\end{subfigure}  
\caption{Training data provided by LiTS-challenge}
			\label{fig:daten}	
\end{center}
\end{figure}

The following approach is to implement a model using two U-Nets $\mathbb{A}_{\theta}$, 
$\mathbb{B}_{\eta}$,  one on top of the other. The combined network architecture is shown in Figure \ref{fig:model}. The inputs for both CNNs are grey-scale images of size $512 \times 512 \times 1$ and their outputs are binary images of size  $512 \times 512$. While the input of the first U-Net is of the form displayed in Figure~\ref{fig:daten}, the input of the second U-Net is produced by the output of the first one as explained in Section~\ref{sec:general}. 

In both networks, the input passes through an initial convolution layer and is then processed by a sequence of convolution blocks at decreasing resolutions (contracting path). The expanding path of the U-Net then reverses this downsampling process. Skip connections between down- and upsampling path intend to provide local information to the global information while upsampling. As final step the output of the network  is passed to a linear classifier that outputs (via sigmoid) a probability for each pixel being within the liver/tumor.
The model is implemented in Keras\footnote{https://keras.io/} with the TensorFlow backend\footnote{https://www.tensorflow.org/}.


\subsection{Mathematical Modelling} \label{sec:general}

In the following, let $\{X_1, \dots , X_N\}\subseteq (\mathbb{R}^{512 \times 512})^N$ and $\left\{Y_1, \dots , Y_N\right\}  \subseteq (\{0, 1 , 2\}^{512 \times 512})^{N}$ denote the 
set of training images and the  corresponding segmented  images, respectively.   
Here the label $1$ stand for liver, $2$ for tumor and $0$ for  background.    
For the task of semantic liver and tumor segmentation, we generate  
segmentation masks 
\begin{align*}
\{A_1, \dots , A_N \}  &\subseteq \big(\{0,1\}^{512 \times 512}\big)^{N} \\
\{B_1, \dots , B_N \}  &\subseteq \big(\{0,1\}^{512 \times 512}\big)^{N}
\end{align*} 
representing binary images $A_k$  where class label 1 
stands for  the  liver or tumor,
and binary images $B_k$  where class label $1$ stands  for 
tumor. 

Our approach is to  train  two networks   
\begin{align*}
\mathbb{A}_{\theta} \colon \mathbb{R}^{512 \times 512} 
 \rightarrow \left[0,1\right]^{512 \times 512}
\\
\mathbb{B}_{\eta} \colon \mathbb{R}^{512 \times 512} \rightarrow 
\left[0,1\right]^{512 \times 512}   
\end{align*}
that  separately   perform  liver and tumor  segmentation.     
In the fist step, the network   $\mathbb{A}_{\theta}$
is applied such that  $\mathbb{A}_{\theta}(X_k) \simeq A_k$. 
After decision making by selecting a threshold $t_a \in (0,1)$, we obtain a liver mask 
$\mathbb{M}_{\theta} \colon \mathbb{R}^{512 \times 512} \rightarrow 
\{0,1\}^{512 \times 512}$ that is applied to  each 
input  image.  Additionally, 
we applied windowing  $w$ pointwise to  the intensity values, 
which results in   new training data 
\begin{align*}
\begin{split}
\bar{X}_{k}  &= w \big(\mathbb{M}_{\theta} (X_k)   X_k\big)
\\
\bar{B}_{k} &= \mathbb{M}_{\theta} (X_k)  B_k \,.
\end{split}
\end{align*}
These data   serve  as input and corresponding ground truth 
for  training the second network 
$\mathbb{B}_{\eta}$. By selecting  another threshold, a mask  
 $\mathbb{T}_{\eta}$  for the tumors is given.

The final classification  can be performed in assigning  a pixel  $(i,j)$ to class label $2$ 
if $\mathbb{M}_{\theta} = \mathbb{T}_{\eta} =1$, to class label $1$ 
if $\mathbb{M}_{\theta} = 1$ and $\mathbb{T}_{\eta} =0$, and class label 0 otherwise.  
The goal is  to find the high dimensional parameter vectors  
 $\theta$ and $\eta$ such that the overall classification error is small.
 This is achieved  by minimizing  a loss function that describes how well the network 
 performs on the training data. Here we  propose to use the joint 
 loss function  
\begin{multline}\label{eq:jointloss}
\mathcal{L}(\theta, \eta) = 
  \frac{c}{N} \sum_{k=1}^N     L\big(\mathbb{A}_{\theta}(X_k), A_k\big) 
 + \\
\frac{1-c}{N} \sum_{k=1}^N  L\big(\mathbb{B}_{\eta}(w \big(\mathbb{M}_{\theta} (X_k)   X_k\big)),  \mathbb{M}_{\theta} (X_k)   B_k \big)  \,,
\end{multline}
where $L$ denotes the categorical cross-entropy-loss and the constant 
$c$ weights the importance of the two classification outcomes. It has been 
demonstrated in \cite{adler2018task} that a joint loss function can  improve results 
compared  to  sequential approaches for joint image reconstruction and segmentation.

\subsection{Optimization of the Models}

The One-Step architecture introduced in ~\ref{sec:one} is pre-trained for 50 epochs applying categorical-cross-entropy loss and fine-tuned for further 30 epochs using the balanced version of the loss stated (see~\ref{alpha}). Balanced loss proved very useful in detecting the lesion for both methods.\\

As already mentioned, the second approach works sequentially, which means that first we optimize $\theta$ and then use the output of $\mathbb{A}_{\theta}$ as input for $\mathbb{B}_{\eta}$. Specifically, for training the second U-Net we minimize  

\begin{align}\label{eqn:loss}
\mathcal{L}_{\mathbb{B}}&(\theta, \eta)=\nonumber\\ &-\frac{1}{N}\sum_{k=1}^N \Big[\sum_{i,j= 1}^{512} \alpha \mathbbm{1}_{\{(a,b)|\bar B_k^{a,b} = 0\}}(i,j) \log \big(\mathbb{B}_{\eta}(X_k)^{i,j}\big)\nonumber \\
 &+ (1- \alpha)\mathbbm{1}_{\{(a,b)|\bar B_k^{a,b} = 1\}}(i,j)  \log \big(1-\mathbb{B}_{\eta}(X_k)^{i,j}\big)\Big]  .
\end{align}

Here $\bar B_k^{i,j}$  is the  value  of $\bar B_k $ at pixel $(i,j)$, and  the  indicator function 
$\mathbbm{1}$ declares whether $(i,j)$  belongs the  the class 
tumor or not. The weight $\alpha \in (0,1)$ controls the relative importance assigned to the two classes. The best results for both models are achieved by applying \textbf{balanced loss}, whose weights have the form
\begin{align}\label{alpha}
\alpha_k = 1-\dfrac{\abs{\{(a,b)|B_k^{a,b}=1\}}}{\abs{B_k}}
\end{align}
for $k \in 1, \dots, N$ for the model described in~\ref{sec:seq}. The coefficients $t_k, L_k$ and $o_k$ computed as 
\begin{align}\label{alpha}
t_k = 1-\dfrac{\abs{\{(a,b)|T_k^{a,b}=1\}}}{\abs{B_k}}\nonumber \\
l_k = 1-\dfrac{\abs{\{(a,b)|L_k^{a,b}=1\}}}{\abs{L_k}}\nonumber \\
o_k = 1-\dfrac{\abs{\{(a,b)|O_k^{a,b}=1\}}}{\abs{O_k}}
\end{align}
denote the balanced class weights for tumor, liver and other tissue respectively, applied in the method depicted in~\ref{sec:one}.

Both models have been trained using stochastic  gradient decent with momentum
for 300 and 600 epochs, respectively. Each iteration takes about 70 seconds on NVIDIA standard GPU. To avoid overfitting, we applied a dropout of $0.4$ in the upsampling path. Both U-Nets were trained with a learning rate of 0.001 and categorical cross-entropy loss. Since the tumor area only accounts for a small area compared to the full size of the image, we applied balanced loss (\ref{eqn:loss}) in a second optimization of the network and reduced the learning rate to 0.0001. 
Comparison  with the   joint loss  \eqref{eq:jointloss} is subject of future    
work.

\section{Experimental Results}


\subsection{Datasets}

The network training is run using a subset of the publicly available LiTS-Challenge\footnote{https://competitions.codalab.org/competitions/17094}dataset containing variable kinds of liver lesions (HCC, metastasis, \dots). The dataset consists of CT scans coming from different clinical institutions. Trained radiologists have manually segmented annotation of the liver and tumors. All of the volumes were enhanced with a contrast agent, imaged in the portal venous phase. Each volume contains a variable number of axial slices with a resolution of $512 \times 512$ pixels and an approximate slice thickness ranging from 0.7 to $\SI{5}{mm}$. The training is applied on 765 axial slices, 50 are used for validation and 50 for testing.

\begin{figure}[htb!] \centering
				\begin{subfigure}[b]{0.8\textwidth}
  	\includegraphics[width=0.48\textwidth]{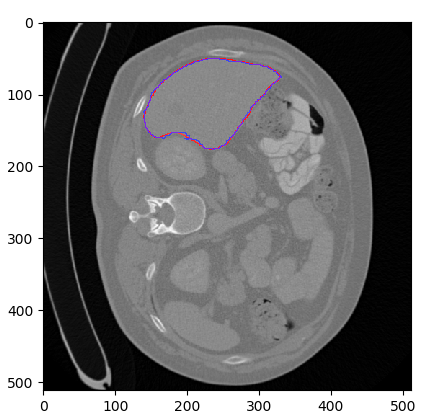}
  	\hfill
  	\includegraphics[width=0.48\textwidth]{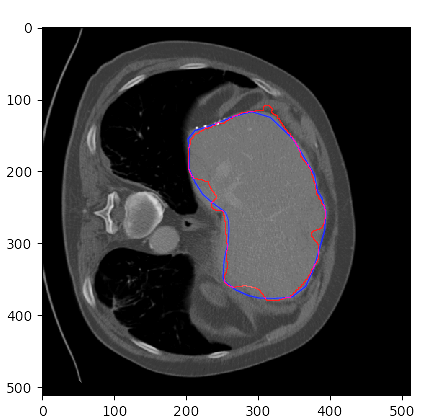}

 	\centering
  	\includegraphics[width=0.48\textwidth]
  	{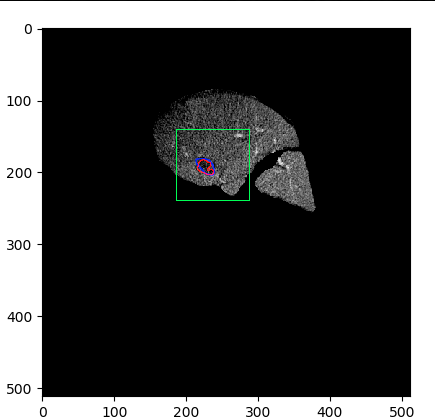}
  	\hfill
  	\includegraphics[width=0.405\textwidth]{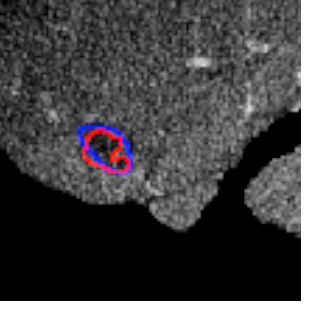}
  		\centering
  	\includegraphics[width=0.46\textwidth]
  	{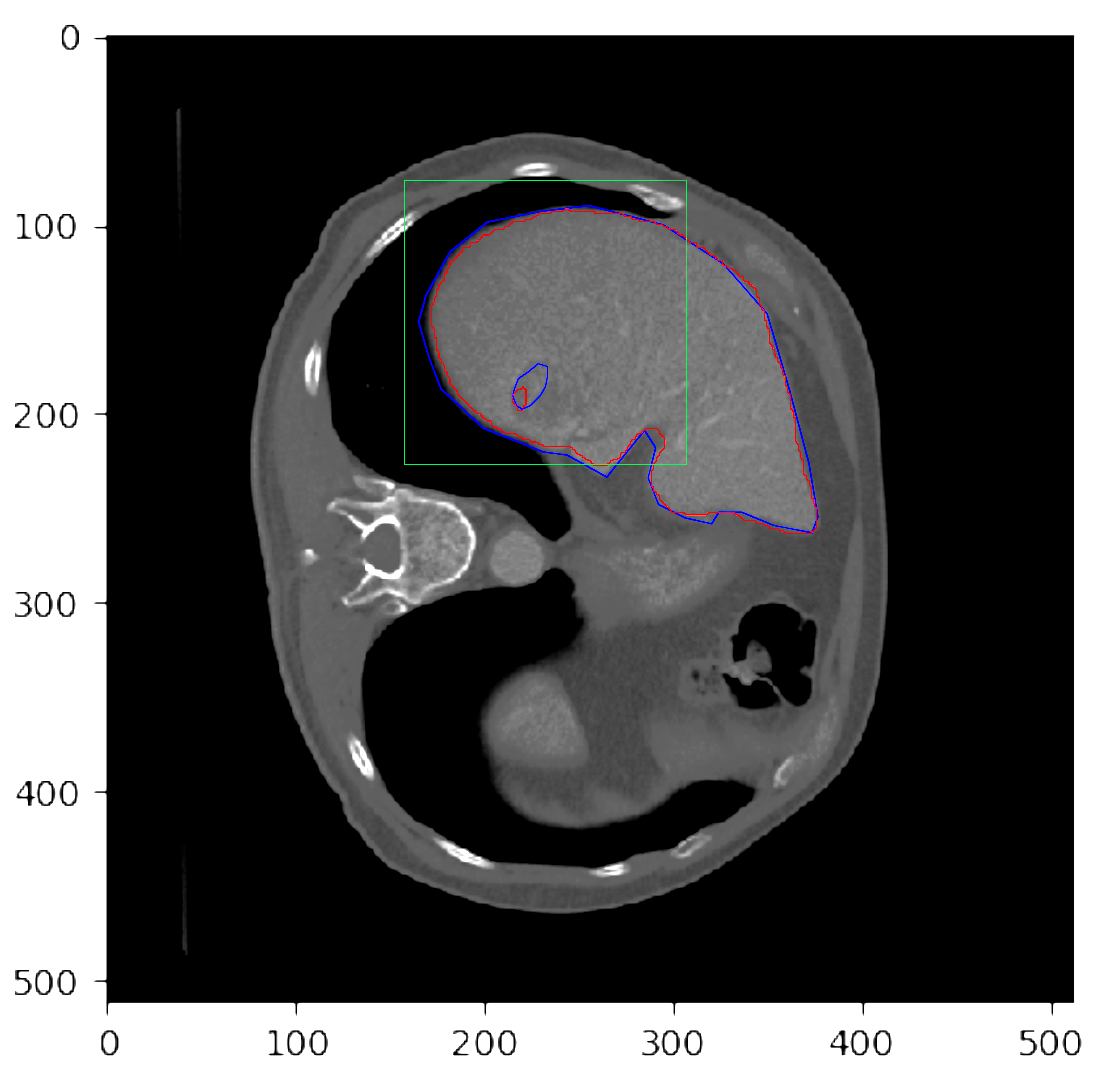}
  	\hfill
  	\centering
  	\includegraphics[width=0.41\textwidth]{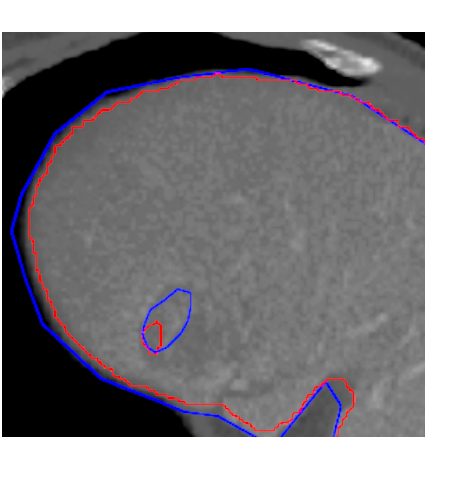} 	
  	\end{subfigure}
		\caption{\textbf{Results on HCC data.} 
		\textbf{Top}: liver segmentation results (red) compared to ground truth boundary (blue). The left image pertains to the LiTS-Challenge dataset, the right one is part of the test set from Innsbruck.
		\textbf{Second row}: tumor segmentation result (red) compared to ground truth (blue) of radiological center in Innsbruck resulting from the sequential approach decribed in~\ref{sec:seq}. \textbf{Bottom}: Segmentation maps preserved by applying ~\ref{sec:one}. }
	 \label{fig:23}
\end{figure}

Further test data is provided by radiological center at the medical university of Innsbruck. The dataset contains CT scans of patients suffering from HCC and the belonging reference annotations were drafted by medical scientists.
Because deep learning algorithms achieve better performance if the data has a consistent scale or distribution, all data are standardized to have intensity values between $[0,1]$
before starting the optimization.

\subsection{Evaluation on Test Data}

 Each pixel of the image is assigned to one of the two classes liver/other tissue and tumor/other tissue, respectively, with a certain probability. Results of the automated liver and tumor segmentation are visualized in Figure~\ref{fig:23}. Comparison with ground truth and segmented liver and tumor give rise to the assumption that our approach is highly promising for obtaining high performance metrics. 
To qualitatively evaluate performance,  we applied some of the commonly used evaluation metrics in semantic image segmentation. 
Performance metrics are summarized in Table~\ref{tab:evalresults}.

\subsubsection{AUC metric} \label{sec:AUC}

Area under ROC Curve (AUC) is a performance metric for binary classification problems. We applied ROC analysis to find the threshold that achieves the best results for the tumor segmentation task. Due to the very low rate of false classified  pixels (most of them has probability close to one or close to zero), we decided to restrict the ROC curve to pixels whose probability for belonging to class tumor lies between 0.01 and 0.99. 

\begin{figure}[htb!] \centering	
  	\includegraphics[width=0.8\textwidth]{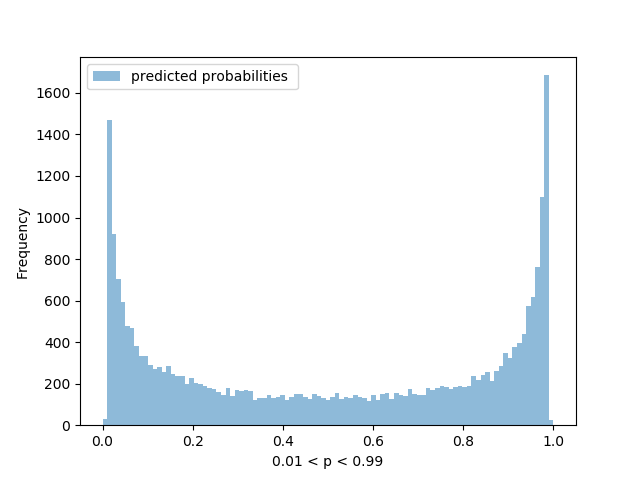}
	\caption{Histogram that displays the number of pixels predicted falling into class tumor with probability $p \in (0.01, 0.99)$ (predictions made by the sequential model).}\label{fig:tumorhist}
\end{figure}

\begin{figure}[htb!]
				\centering
  	\includegraphics[width=0.8\textwidth]{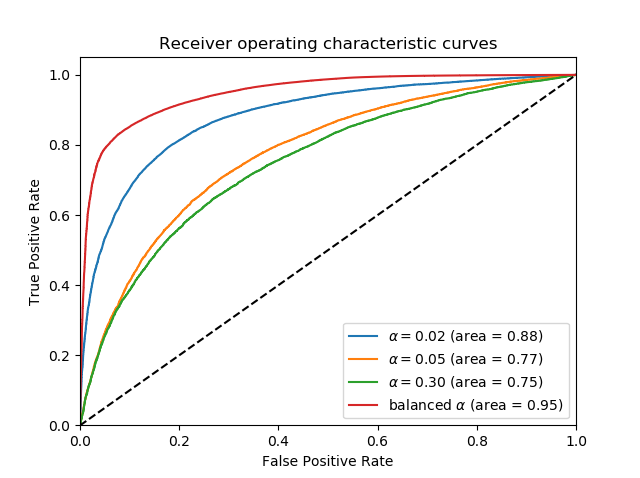}		\caption{Restricted ROC curves for varying weights $\alpha$ of weighted and balanced loss. The red curve corresponds to the outcomes produced by applying balanced loss, which apparently leads to the best tumor segmentation results. In general terms it can be stated that setting the importance of the background pixels lower seems to considerably improve segmentation accuracy of the lesion.}\label{fig:roctum}
 \end{figure}

In Figure~\ref{fig:roctum} we can see that the best restricted AUC value (rAUC) conducting $0.88$ is achieved by a very small value of $\alpha = 0.02$. We further calculated the corresponding threshold and could achieve an improvement of the tumor segmentation results \cite{kumar2011receiver}. 

\subsubsection{Intersection over Union}

For a more complete evaluation of the segmentation results we use class accuracy in conjunction with the so called IoU metric. The latter is essentially a method to quantify the percent overlap between the ground truth and the prediction output. The IoU measure gives the similarity between predicted  and ground-truth regions for the object of interest. The formula for quantifying the IoU score is:
\begin{equation}
\text{IoU} = \frac{\text{TP}}{\text{FP} + \text{TP} + \text{FN}}
\end{equation}
 where \text{TP}, \text{FP} and \text{FN} denote the True Positive Rate, False Positive Rate and False Negative Rate, respectively.

\begin{figure}[htb!]
\centering 
\begin{subfigure}[b]{0.8\textwidth}
  	\includegraphics[width=0.48\textwidth]{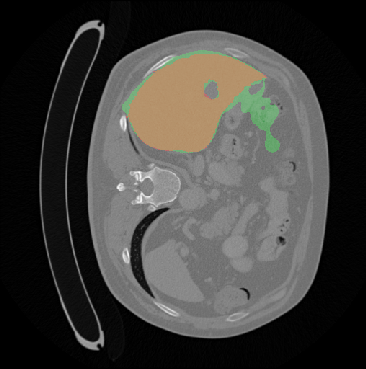} 
  	\hfill
   \includegraphics[width=0.48\textwidth]{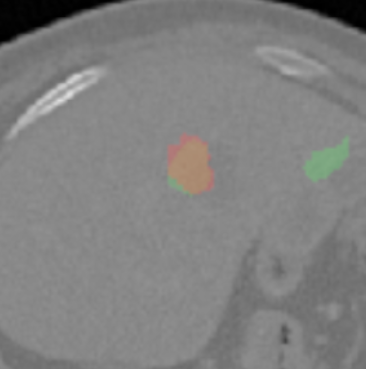}
			\centering
  	\includegraphics[width=0.48\textwidth]{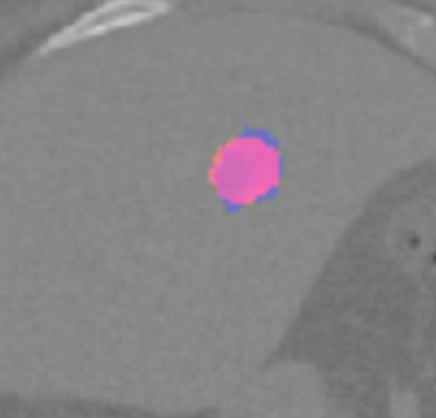} 
  	\hfill
   \includegraphics[width=0.48\textwidth]{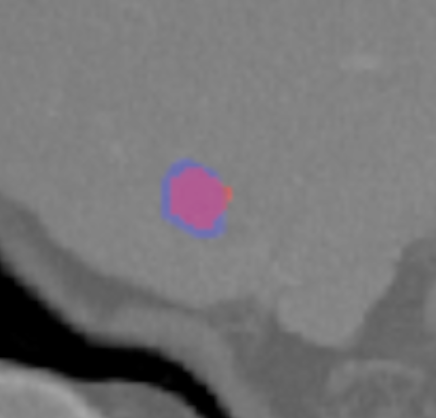}
  
\end{subfigure}
\caption{\textbf{Top:} Intersection over Union of liver and tumor segmentation resulting from the one-step model. The green area indicates the predicted masks, the orange are is their overlap and the share highlighted in red, is the ground truth.  
\textbf{Bottom:} Intersection over Union of tumor segmentation for balanced loss with balanced $\alpha$ resulting from the sequential model. The light pink, light blue and pink areas mark the  prediction mask, ground truth and  Intersection over Union, respectively. }
\end{figure}

Since the segmentation task can be regarded as clustering of pixels, Rand index \cite{jain2010boundary}, which is a measure of the similarity between two data clusterings, has been proposed as a measure of segmentation performance. By $S_1$ and $S_2$ two segmentations are notated. In the following paragraph $S_1$ denotes the ground truth and $S_2$ the segmentation results obtained by the joint network. The function $\delta$ is defined as $\delta(S_{i_j}, S_{i_k})=0$ if  pixels $j$ and $k$ are in same class
and $\delta(S_{i_j}, S_{i_k})=1$ otherwise. One can see that small differences in the location of object boundaries will increase the rand error slightly while merging or splitting of objects lead to a big increase of the Rand error.

We evaluate this model under usage of test and validation set from LiTS-Challenge and Innsbruck data ($112$ images). The evaluation metrics are summarized in Table \ref{tab:evalresults}. The liver segmentation evaluation scores indicate that our models, especially perform remarkable good, provided that the sequential approach outperforms the One-Step method primarily in the tumor segmentation task. in  Pixel accuracy, Intersection over union (IoU) and Rand Index (RI) have values very close to $1$. IoU and Rand Index performance score of the tumor segmentation show that the application of balanced loss with  $\alpha = 0.02$, achieves the best results.  

\begin{center}
 \begin{tabular}{lccccc}
  Data & $\alpha$ & rAUC& $\text{Pixel}_{acc}$ & IoU & RI \\
\clineB{1-6}{2.5}
\textbf{One-Step} & & & & &\\
\clineB{1-6}{2.5}
  Liver & & & 0.9935 & 0.8898 & 0.9316\\
  \hline
Tumor 

	&	bal	& 0.87 &	0.9995 &	0.6782 & 0.8075\\  		
  \clineB{1-6}{2.5}
\textbf{Sequential} & & & & &\\
\clineB{1-6}{2.5}
  Liver & & & 0.9999 & 0.9385 & 0.9628\\
  \hline
Tumor 
  		&	0.02	& 0.88 &	0.9996 &	0.77108 & 0.8706\\
  		&	0.05&	0.77 &0.9996	&	0.7261 & 0.8490\\
  		&	0.30	& 0.75 &	0.9995	&	0.73879 & 0.8433\\
	&	bal	& \fbox{0.95} &	\fbox{0.9997} &	\fbox{0.7917} & \fbox{0.9106}\\  		
  \hline

 \end{tabular}
  \captionof{table}{Performance Evaluation metrics for tumor segmentation models applied on test data (112 images)}
 \label{tab:evalresults}\end{center}

\section{Conclusions}

We presented two deep learning frameworks for the automated joint liver and tumor segmentation. Segmentation metrics evaluate the segmentation of detected lesions and are comprised of a restricted AUC, an overlap Index (IoU) and Rand Index (RI). The first model manages segmentation of tumor and lesion in one step and works very convenient and fast but, as becomes clear from visualizations and evaluation metric scores, it is prone to misclassification, especially in the tumor segmentation task. The second model proposed in this paper works sequentially and clearly outperforms the one-step method.   Another interesting topic to address is the classification of the tumors detected by a deep learning algorithm.


\end{document}